\documentclass[12pt]{article}

\usepackage{url}
\usepackage{setspace}
\usepackage{graphicx}
\usepackage{multirow}

\usepackage{amsmath} 
\usepackage{amsfonts}

\usepackage{hyperref}
\hypersetup{
	colorlinks=true,
	linkcolor=blue,
	filecolor=blue,      
	urlcolor=blue,
	citecolor=blue,
}

\usepackage[a4paper, total={6in, 8in}]{geometry}

\usepackage[utf8]{inputenc}

\usepackage[super,sort,comma]{natbib}

\begin{document}
	\begin{center}	
\sf {\Large {\bfseries A Physics-Informed Deep Learning Model for MRI Brain Motion Correction}} \\  
		\vspace*{10mm}
		Mojtaba Safari, Shansong Wang, Zach Eidex, Richard Qiu, Chih-Wei Chang, David S. Yu, Xiaofeng Yang,  PhD$ ^{\ddagger} $ \\
	{Department of Radiation Oncology and Winship Cancer Institute, Emory University, Atlanta, GA 30322, United States of America\\$ ^\ddagger  $Corresponding Author}
	\vspace{5mm}\\
\end{center}

Author to whom correspondence should be addressed. email: \url{xiaofeng.yang@emory.edu} \\

\newpage

\begin{abstract}
	\noindent \textbf{Background:} Magnetic resonance imaging (MRI) is an essential brain imaging tool, but its long acquisition times make it highly susceptible to motion artifacts that can degrade diagnostic quality.\\
	\noindent \textbf{Purpose:} This work aims to develop and evaluate a novel physics-informed motion correction network, termed PI-MoCoNet, which leverages complementary information from both the spatial and \textit{k}-space domains. The primary goal is to robustly remove motion artifacts from high-resolution brain MRI images without explicit motion parameter estimation, thereby preserving image fidelity and enhancing diagnostic reliability.\\ 
	
	\noindent\textbf{Materials and Methods:} PI-MoCoNet is designed as a dual-network framework consisting of a motion detection network and a motion correction network. The motion detection network employs a U-net architecture to identify corrupted \textit{k}-space lines using a spatial averaging module, thereby reducing prediction uncertainty. The correction network, inspired by recent advances in U-net architectures and incorporating Swin Transformer blocks, reconstructs motion-corrected images by leveraging three loss components: the reconstruction loss ($\mathcal{L}_1$), a learned perceptual image patch similarity (LPIPS) loss, and a data consistency loss ($\mathcal{L}_{\text{dc}}$) that enforces fidelity in the \textit{k}-space domain. Realistic motion artifacts were simulated by perturbing phase encoding lines with random rigid transformations. The method was evaluated on two public datasets (IXI and MR-ART). Comparative assessments were made against baseline models, including Pix2Pix GAN, CycleGAN, and a conventional U-net, using quantitative metrics such as peak signal-to-noise ratio(PSNR), structural similarity index measure  (SSIM), and normalized mean square error (NMSE).\\	
	
	\noindent\textbf{Results:} 	PI-MoCoNet demonstrated significant improvements over competing methods across all levels of motion artifacts. On the IXI dataset, for minor motion artifacts, PSNR improved from 34.15 dB in the motion-corrupted images to 45.95 dB after correction, SSIM increased from 0.87 to 1.00, and NMSE was reduced from 0.55\% to 0.04\%. For moderate artifacts, PSNR increased from 30.23 dB to 42.16 dB, SSIM from 0.80 to 0.99, and NMSE from 1.32\% to 0.09\%. In the case of heavy artifacts, PSNR improved from 27.99 dB to 36.01 dB, SSIM from 0.75 to 0.97, and NMSE decreased from 2.21\% to 0.36\%. On the MR-ART dataset, PSNR values increased from 23.15  dB to 33.01 dB for low artifact levels and from 21.23 dB to 31.72 dB for high artifact levels; concurrently, SSIM improved from 0.72 to 0.87 and from 0.63 to 0.83, while NMSE decreased from 10.08\% to 6.24\% and from 14.77\% to 8.32\%, respectively. An ablation study further confirmed that incorporating both data consistency and perceptual losses led to an approximate 1 dB gain in PSNR and a reduction of 0.17\% in NMSE compared to using the reconstruction loss alone.\\

	\noindent\textbf{Conclusions:} PI-MoCoNet is a robust, physics-informed framework for mitigating brain motion artifacts in MRI. It successfully integrates spatial and \textit{k}-space information to enhance image quality. Its superior performance over comparative methods highlights its potential for clinical application, particularly in settings where patient motion is unavoidable. The source code is available at: \url{https://github.com/mosaf/PI-MoCoNet.git}.
	
\end{abstract}

\textbf{\textit{keywords:}} \textit{k}-space, MRI, motion correction, MoCo, physics informed deep learning, deep learning

\newpage 
\section{Introduction}

Motion artifacts are among the most common and challenging distortions in high-resolution brain Magnetic resonance imaging (MRI) images, often arising from both involuntary and voluntary patient movements~\cite{Sreekumari217}. MRI is a key modality for generating functional and anatomical images that inform diagnosis, treatment, and prognosis; for example, high-resolution brain anatomical MRI is used extensively to delineate tumor subregions and monitor post-treatment outcomes~\cite{Villanueva_Meyer2024_Lancet, sharma2024doestimeretreatmentmatter, Beigi2018_radiologia, Beigi2018}. However, the inherently long acquisition times of MRI greatly increase the likelihood of patient motion, which disrupts spin history, leads to signal loss, and alters the B$_0$ field, leading to susceptibility artifacts~\cite{https://doi.org/10.1002/acm2.14072}. Severe motion also causes inconsistencies in \textit{k}-space, potentially violating the Nyquist criterion~\cite{https://doi.org/10.1002/jmri.24850}, which causes image ghosting and blurring.

Deep learning (DL) algorithms have achieved remarkable success in medical imaging and treatment~\cite{9363915}. Studies have proposed both supervised and unsupervised DL approaches to remove motion artifacts~\cite{Zhou2024}. These methods are trained in the image domain to learn a mapping from motion-corrupted to motion-free space by training a network $f_\nu: \mathcal{T}_a \to \mathcal{T} $, where $\mathcal{T}_a$ and $\mathcal{T}$ are motion-corrupted and motion-free spaces, respectively~\cite{DUFFY2021117756, Safari_2024_maudgan, Liu2021}. While these techniques can effectively recover motion-free images, their exclusive reliance on the image domain may lead to image hallucinations, particularly under severe motion conditions~\cite{https://doi.org/10.1002/mp.16844}. In contrast, several studies have attempted to correct motion artifacts by estimating rigid motion parameters (e.g., translation and rotation) and applying corrections using compressed sensing MRI algorithms~\cite{https://doi.org/10.1002/mp.16119, 10639524}. Although these methods leverage the localized nature of motion artifacts in \textit{k}-space, they can be lengthy and limited to raw \textit{k}-space data that are not available to end users. In addition, the spatial domain representing images provides complementary information that can improve the MoCo models, which might be overlooked by these studies.

In this work, we propose a physics-informed motion correction network (PI-MoCoNet) that leverages both spatial and \textit{k}-space domain information to remove brain motion artifacts. Our approach avoids explicit motion parameter estimation, enabling potential extension to non-rigid motions such as coherent lung motion. Our key contributions are as follows:

\begin{enumerate} 
	\item We develop a realistic motion artifact simulation framework that mimics the actual motion-induced distortions observed in clinical MRI. 
	
	\item We design a novel variation of the U-net architecture inspired by Yue et al.~\cite{10681246}, in which traditional attention layers are replaced by Swin Transformer blocks to enhance feature representation. 
	
	\item Our proposed PI-MoCoNet exploits \textit{k}-space data to predict corrupted lines and enforces data consistency, thereby preserving data fidelity during motion correction. 
	
\end{enumerate}

\section{Materials and Methods}\label{sec:matrial}
\subsection{Problem formulation}
The spine echo MRI acquisition signal for each phase encoding line of a sample can be written as follows:

\begin{equation}
	\centering
	s(k_x, k_y) = \int \int \rho(x, y) e^{-i2\pi (k_xx + k_y y)} = \mathcal{F}[\rho(x, y)]
	\label{eq:spin_echo}
\end{equation}
where $s \in \mathbb{C} ^{N_x \times N_y}$ represents the measured \textit{k}-space data, $k_x\in \mathbb{R}^{N_x}$ and $k_y\in \mathbb{R}^{N_y}$ are the frequency and phase encoding directions, and $\rho \in \mathbb{C} ^{N_x \times N_y}$ encapsulates the combined effects of the receive field, spin density, and the gain of the MRI system. Here, $\mathcal{F}$ denotes the Fourier transform~\cite{doi:https://doi.org/10.1002/9781118633953.ch10}.


During patient motion, some phase encoding lines (i.e., $k_y$ lines) are displaced. These displacements in \textit{k}-space induce artifacts in the reconstructed images, manifesting as blurring, ghosting, or ringing, depending on the location of the affected lines~\cite{9025260}. Motivated by the localized nature of motion artifacts in \textit{k}-space, we hypothesize that enforcing data consistency (DC) of the corrupted \textit{k}-space lines can effectively restore image fidelity.


\subsection{PI-MoCoNet}

\begin{figure}
	\centering
	\includegraphics[width=\textwidth]{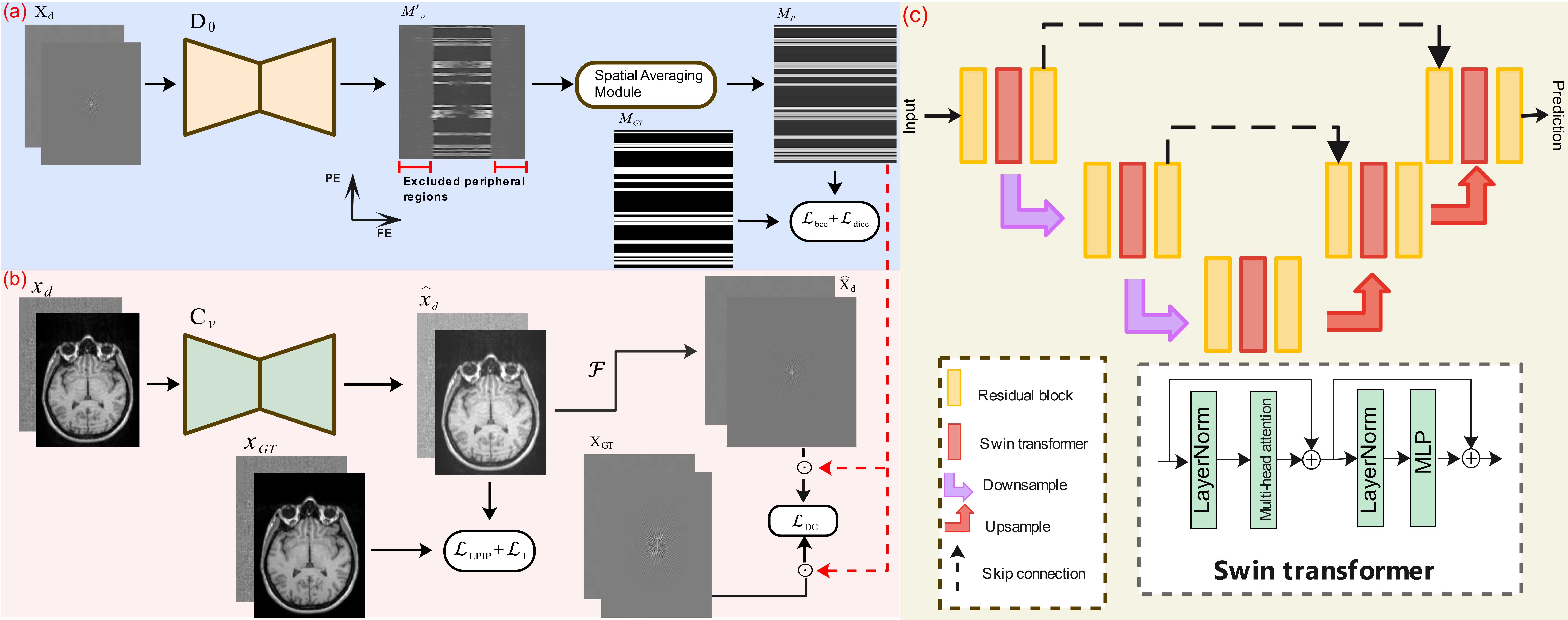}
	\caption{Flowchart of the proposed PI-MoCoNet framework. (a) The motion detection network $\mathcal{D}_\theta$ identifies corrupted regions in the frequency domain, producing a predicted motion mask $M_p$ after spatial averaging. The peripheral regions are excluded to enhance accuracy.	(b) The motion correction network $\mathcal{C}_\nu$ refines the input motion-corrupted images by learning a mapping to the ground truth, minimizing perceptual and reconstruction losses ($\mathcal{L}_{lpips} + \mathcal{L}_1$). The corrected image is further evaluated using a DC loss, incorporating a predicted corruption mask ($M_{GT}$, indicated by the red dashed line). (c) The artifact removal network adopts a U-net architecture enhanced with Swin Transformer blocks for improved feature extraction. Residual blocks, downsampling, and upsampling operations are integrated, with skip connections preserving spatial information. The lower and uppercase $x$ and $X$ represent images in the spatial and frequency domains, respectively, while $\odot$ denotes pixel-wise multiplication.}
	
	\label{fig:flowchart_01}
\end{figure}

Our proposed PI-MoCoNet consists of two interconnected networks: a motion detection network $\mathcal{D}_\theta$ and a motion correction network $\mathcal{C}_\nu$. These networks are trained simultaneously to detect the corrupted \textit{k}-space lines and correct them, respectively. \figurename~\ref{fig:flowchart_01} provides an overview of the proposed framework.

\subsection{Motion detection network $\mathcal{D}_\theta$}

In the motion detection network ($\mathcal{D}_\theta$) illustrated in \figurename~\ref{fig:flowchart_01}a, a U-net architecture~\cite{10643318} is employed to identify motion-corrupted regions directly in the \textit{k}-space domain. To enhance prediction reliability, particularly in low-SNR regions at the peripheral areas of \textit{k}-space, we introduced a ``spatial averaging module'' defined as follows:

\begin{equation}
	\centering
	M (x, y) = \frac{1}{N_y} \sum_{y^\prime} M^\prime (x, y^\prime),  \,\,\,\, \text{for all}\,\,\,\, y
\end{equation} 
where $ M^\prime $ is the first prediction map of the motion detection network $ \mathcal{D}_\theta $. The spatial averaging module computes the average prediction values of the $ \mathcal{D}_\theta $ across the frequency encoding direction and assigns these average values to the corresponding \textit{k}-space lines. This step reduces the prediction uncertainty and ensures consistency with the phase encoding direction.

The network is trained to minimize a combination of the Dice loss and binary cross-entropy (BCE) loss between the predicted mask $M_p$ and the ground truth mask $M_{GT}$:

\begin{equation}
	\centering
	\mathcal{L}_{\text{seg}} (\theta) = \mathcal{L}_{\text{Dice}}(M_p, M_{GT}) + \mathcal{L}_{\text{BCE}}(M_p, M_{GT})
	\label{eq:loss_seq}
\end{equation}
where  

\begin{equation}
	\mathcal{L}_{\text{Dice}}(M_p, M_{GT}) = 1 - \frac{2 \sum M_p M_{GT} + \epsilon}{\sum M_p + \sum M_{GT} + \epsilon}
\end{equation}

\begin{equation}
	\mathcal{L}_{\text{BCE}}(M_p, M_{GT}) = - \sum \left[ M_{GT} \log M_p + (1 - M_{GT}) \log (1 - M_p) \right]
\end{equation}
and $\epsilon = 1\times 10^{-6}$ is a small constant to prevent division by zero.

\subsection{Motion correction network $\mathcal{C}_\nu$}

The majority of the studies utilized U-net architecture constructed using convolution layers, which may perform better in learning high-frequency representations compared to transformers~\cite{Li_ASPS_MICCAI2024, 10852524}. However, their small receptive field limits them to extract local representations. To address this, attention mechanisms, such as attention gates and squeeze-and-excitation blocks, have been integrated into U-net variants to enhance feature selection and improve segmentation performance. Conversely, transformers are able to extract long-range dependencies~\cite{https://doi.org/10.1002/mp.17600}.

In this study, drawing inspiration from Yue \textit{et al.}~\cite{10681246}, we employed a U-net architecture that substitutes the attention layers with Swin Transformer blocks. This approach was chosen due to its superior ability to generalize across various image resolutions in image restoration tasks such as image denoising and super-resolution~\cite{10681246, 9607618}. Our proposed motion correction network, denoted as $\mathcal{C}_\nu$, alongside the U-net backbone network, is illustrated in \figurename~\ref{fig:flowchart_01}b and c, respectively.

Our motion correction network $\mathcal{C}_\nu$ aimed to recover the motion-corrected images $\hat{x}_d$. We used two losses in the image space, including the $\mathcal{L}_1$ loss to preserve data fidelity and the learned perceptual image patch similarity (LPIPS) loss $\mathcal{L}_{{lpips}}$~\cite{8578166} to maintain perceptual similarity between motion-corrected $\hat{x}_d$ and ground truth $x_{GT}$ images. Additionally, the data consistency loss $\mathcal{L}_{\text{dc}}$ in \textit{k}-space was also used to enforce consistency between the images due to the localized distortion in \textit{k}-space. The motion correction network losses are formulated as follows:

\begin{equation}
	\mathcal{L}(\nu) = \lambda_r \mathcal{L}_1(\hat{x}_d, x_{GT}) + \lambda_{{l}} \mathcal{L}_{{lpips}}(\hat{x}_d, x_{GT}) + \lambda_{{d}} \mathcal{L}_{\text{dc}}(\hat{x}_d, x_{GT}),
\end{equation}

where  

\begin{equation}
	\mathcal{L}_1(\hat{x}_d, x_{GT}) = \|\hat{x}_d - x_{GT} \|_1
\end{equation}

is the $\mathcal{L}_1$ loss,  

\begin{equation}
	\mathcal{L}_{{lpips}}(\hat{x}_d, x_{GT}) = \sum_i \| \phi_i(\hat{x}_d) - \phi_i(x_{GT}) \|_2^2
\end{equation}

is the LPIPS loss, where $\phi_i$ represents the activation maps extracted from a pre-trained deep network, and  

\begin{equation}
	\mathcal{L}_{\text{dc}}(\hat{x}_d, x_{GT}) = \| \mathcal{F}(\hat{x}_d) \odot M_p - \mathcal{F}(x_{GT}) \odot M_p \|_2^2
\end{equation}

is the data consistency loss in \textit{k}-space, where $\odot$ denotes the pixel-wise multiplication. The hyperparameters $\lambda_r = 10$, $\lambda_{{l}} = 0.5$, and $\lambda_{{d}} = 100$ are weighting factors to balance the contribution of each loss term.

\subsection{Motion simulation}

Acquiring a large dataset of motion-corrupted and motion-free image pairs for training DL models is challenging. To overcome this challenge, we generated an \textit{in-silico} dataset by simulating realistic motion artifacts. Specifically, we assumed abrupt rigid brain motions occurring at a frequency faster than the frequency encoding sampling rate but slower than the phase encoding sampling rate. Consequently, realistic brain motion artifacts were simulated by selectively modifying \textit{k}-space lines along the phase encoding (PE) direction.

Given a motion-free \textit{k}-space $ k_{\text{GT}} \in \mathbb{C} ^{N \times N}$, random translation and rotation parameters, denoted as $\Theta_i \in \mathbb{R}^3$, were sampled to perform a rigid transformation using the generated motion trajectories. Subsequently, a corruption mask MM was randomly generated by selecting k-space slabs of varying widths along the PE direction. These masks were then applied to replace the corresponding lines of the original \textit{k}-space data with their motion-corrupted counterparts, as illustrated in \figurename~\ref{fig:motion_simulation_v1}. The motion simulation process can be mathematically expressed as:

\begin{figure}
	\centering
	\includegraphics[width=\textwidth]{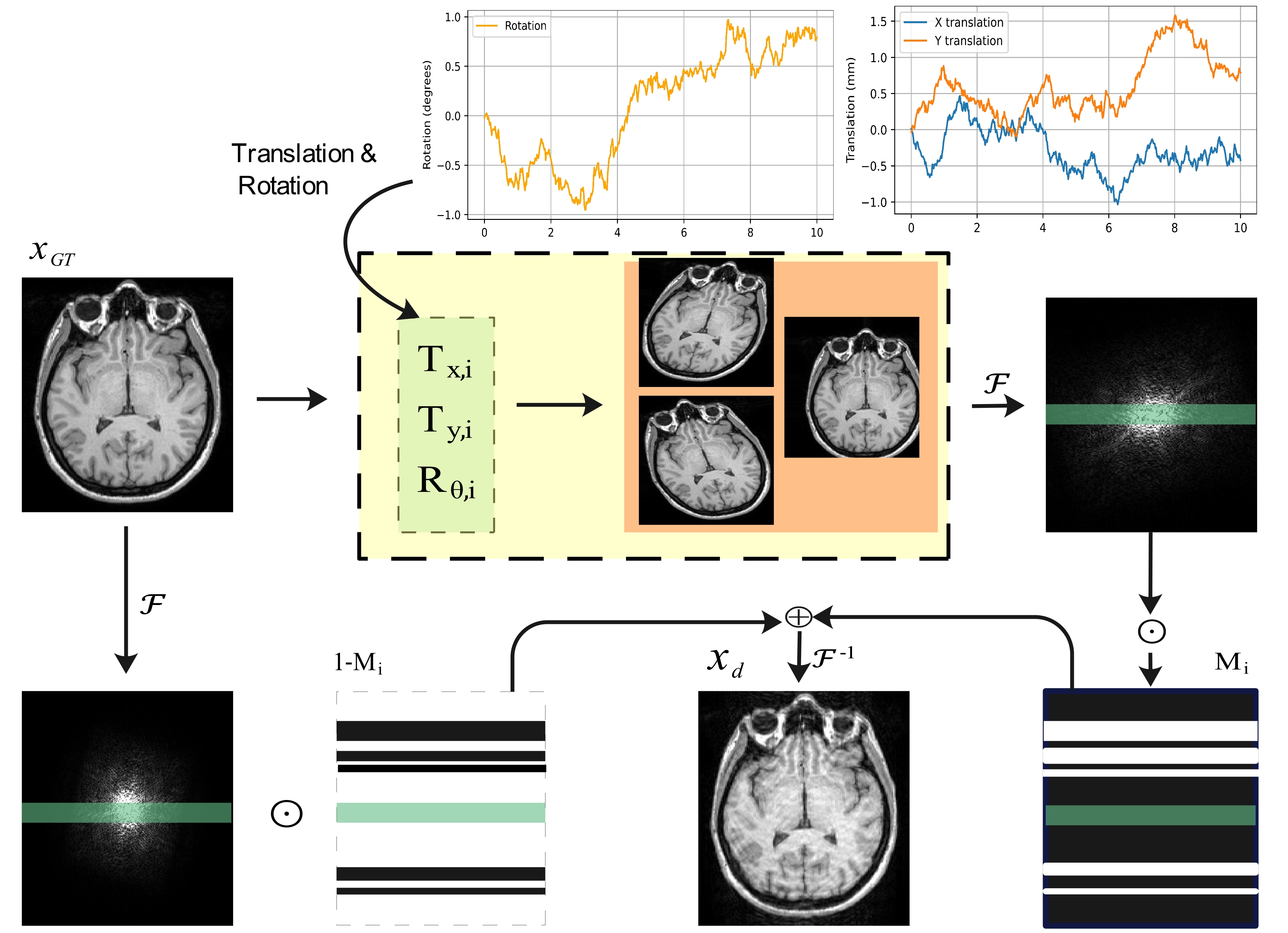}
	
	\caption{Motion simulation process. A motion-free input image undergoes a transformation using randomly sampled translation ($T_{x,i}, T_{y,i} $) and rotation  parameters. The transformed image is then converted to \textit{k}-space via the Fourier transform $\mathcal{F}$. Specific \textit{k}-space slabs along the phase encoding (PE) direction are selected, with the mask $M_i$ determining which \textit{k}-space lines are replaced by the motion-corrupted data. This process is repeated $N$ times to achieve the desired level of motion artifacts. The green slab, corresponding to the center of \textit{k}-space, remains excluded from the motion simulation process to preserve image fidelity.}
	\label{fig:motion_simulation_v1}
	
\end{figure}

\begin{equation}
	\centering
	k_{\text{motion}} = \sum_{i=1}^{N} (1 - M) \odot k_{\text{GT}} + M \odot \mathcal{T}_{\Theta_i} (k_{\text{GT}})
\end{equation}
where $N$ denotes the number of discrete motion events and $ \mathcal{T}_{\Theta_i} : \mathbb{R}^{N \times N} \to \mathbb{R}^{N \times N} $ denotes the rigid transformation which is defined as below:

\begin{equation}
	\centering
	\mathcal{T}_{\Theta_j} (k) =\mathcal{F} \circ R_{\theta_i} \circ T_{x_i,y_i} \circ \mathcal{F}^{-1}(k)
	\label{eq:transformation}
\end{equation} 

Here, $\mathcal{F} $ and $\mathcal{F}^{-1}$ denote the Fourier and inverse Fourier transforms, respectively, while $R_{\theta_i}$ and $ T_{x_i,y_i} $ represent the rotation and translation operations in the image domain. The corruption mask $M$ selectively replaces specific \textit{k}-space lines in the PE direction with their motion-corrupted counterparts, ensuring a realistic simulation of brain MRI motion artifacts.

The PI-MoCoNet was implemented using the \texttt{PyTorch} (version 2.5.1)~\cite{paszke2019pytorchimperativestylehighperformance} deep learning framework and executed on an NVIDIA A100 GPU. The model was trained with a batch size of $32$ and a learning rate of $2\times 10 ^{-4}$. The training was conducted for $25$ epochs using the Adam optimizer, with hyperparameters set to $\alpha=0.9$ and $\beta=0.999$.

\subsection{Dataset}\label{sec:data}

We utilized two publicly available datasets, IXI (\url{https://brain-development.org/ixi-dataset/}) and the movement-related artifacts (MR-ART) dataset from OpenNeuro, to train and evaluate our models~\cite{Narai2022_mrART}.

The IXI dataset is comprised of 580 cases with T1-weighted brain MRI images. The dataset was partitioned into two non-overlapping subsets: a training set (n=480, 54,160 slices) and a testing set (n=100, 11,980 slices). To simulate different levels of motion artifacts, three levels of corruption—high, moderate, and minor—were introduced by perturbing 15, 10, and 5 \textit{k}-space slabs, respectively. Random slabs were uniformly sampled, containing between 3 to 7 \textit{k}-space lines, and were affected by rotation artifacts of $\pm 7^\circ$ and translation artifacts of $\pm 5 $ mm.

To assess model performance on \textit{in vivo} images, we employed the MR-ART dataset, which consists of 148 cases (95 females and 53 males). This dataset includes three types of images: motion-free ground truth images, motion-corrupted images at level 1, and motion-corrupted images at level 2, where level 2 represents a higher degree of motion artifacts than level 1.

Institutional Review Board approval was not required for this study, as both datasets were obtained from open-access repositories, and the original studies had already received ethical approval.

\subsection{Quantitative and Statistical Analysis}
We evaluated our proposed method against three benchmark models: U-net with residual connections, CycleGAN, and Pix2Pix GAN. All comparative models were trained using identical datasets to ensure a fair comparison.

Performance was assessed using three quantitative metrics: normalized mean square error (NMSE), peak signal-to-noise ratio (PSNR), and structural similarity index (SSIM)~\cite{1284395}. These metrics were calculated utilizing the PyTorch Image Quality Library~\cite{kastryulin2022pytorchimagequalitymetrics}. Higher values of SSIM and PSNR indicate superior reconstruction quality, whereas lower NMSE values denote better performance. While NMSE may favor the generation of blurrier images, PSNR's logarithmic scale aligns more closely with human perceptual judgments~\cite{safari2024selfsupervisedadversarialdiffusionmodels}. Additionally, SSIM measures the structural similarity between reconstructed and ground truth images, providing insight into the preservation of image features.

To statistically compare the quantitative metrics across different methods, a one-way analysis of variance (ANOVA) was conducted to evaluate the null hypothesis that the mean values of each method are the same. Following ANOVA, Tukey's honestly significant difference (HSD) test was used for comparisons between methods. Differences were considered statistically significant at {p}-value of $< 0.05$.

The average values of the quantitative metrics are presented along with their 95\% confidence intervals (CIs), calculated using the percentile bootstrap method with 10,000 iterations and the bias-adjusted accelerated bootstrap technique. All statistical analyses were performed using \texttt{statsmodels} (version 0.14.4) Python package~\cite{seabold2010statsmodels}.

\section{Results}

This section presents both qualitative and quantitative results obtained from the \textit{in-silico} and \textit{in-vivo} datasets. In addition, we report ablation study findings to evaluate the contributions of each component in the proposed PI-MoCoNet model.

\subsection{Qualitative results}
\begin{figure}
	\centering
	\includegraphics[width=\textwidth]{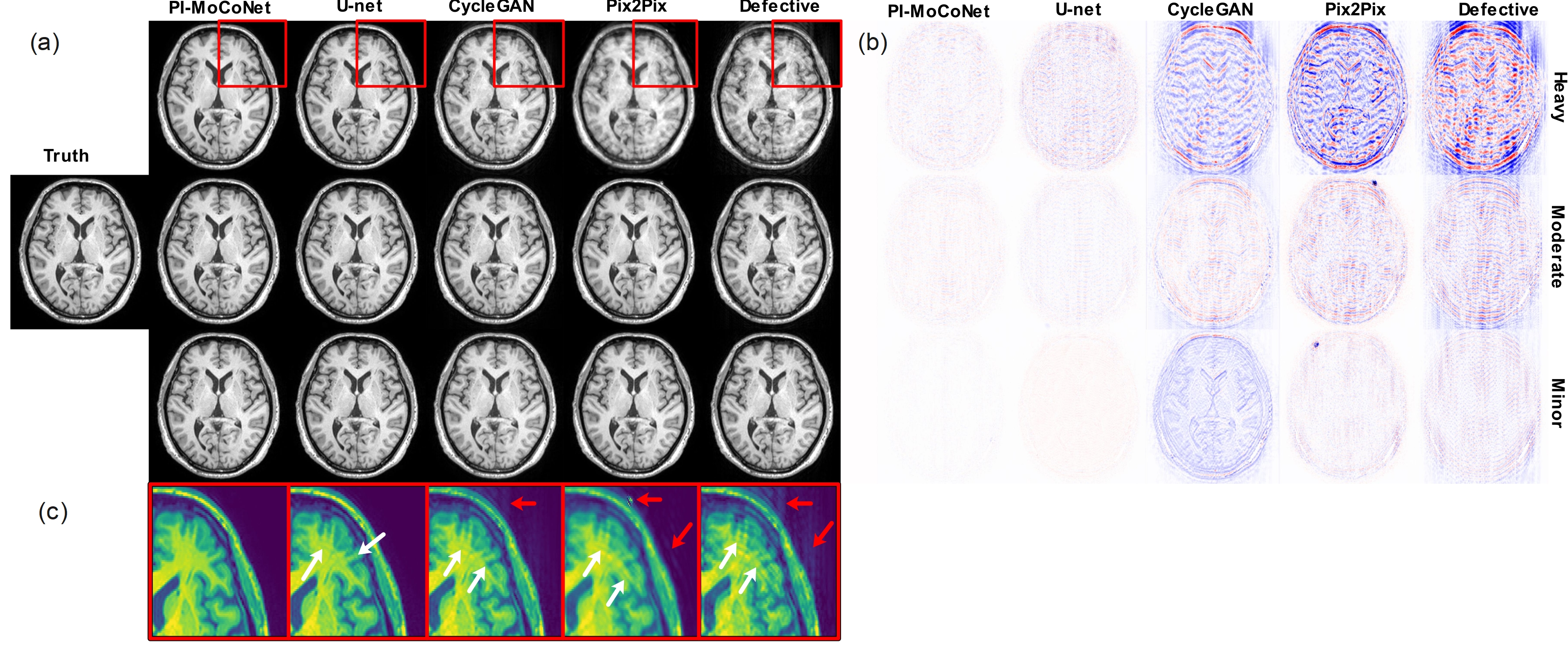}
	\caption{Qualitative evaluation of various methods for removing brain MRI motion artifacts at different severity levels (heavy, moderate, and minor). (a) Reconstruction outcomes, (b) corresponding difference maps, and (c) zoomed-in views of selected regions. White arrows point to remaining ringing artifacts inside the skull, and red arrows highlight ghosting of bright fat tissue outside of the skull.}
	\label{fig:qualitative_figure}
\end{figure}

Our motion simulation approach could successfully simulate ringing artifacts inside the skull and ghosting of bright fat tissue outside of the skull, as shown by white and red arrows in \figurename~\ref{fig:qualitative_figure}, for three different motion artifact levels. 
While both CycleGAN and Pix2Pix effectively recover motion-free images from mildly and moderately corrupted data, they exhibit decreased performance in the presence of severe motion artifacts, as evidenced by residual ringing (white arrows) and ghosting (red arrows) in \figurename~\ref{fig:qualitative_figure}c. This observation is further supported by the difference maps shown in \figurename~\ref{fig:qualitative_figure}b, which indicate droplet artifacts in Pix2Pix GAN and a bias error in CycleGAN. In contrast, U-net successfully removes low and moderate motion artifacts and demonstrates better performance than CycleGAN and Pix2Pix in handling severe artifacts; however, some residual ringing and ghosting remain, as seen in \figurename~\ref{fig:qualitative_figure}c.

Notably, the proposed PI-MoCoNet achieves consistent artifact removal across all severity levels, yielding virtually no remaining ringing or ghosting. The superior performance of PI-MoCoNet is further corroborated by the difference maps (\figurename~\ref{fig:qualitative_figure}b), which indicate minimal discrepancies relative to the ground truth when compared with the other examined methods.

\subsection{Quantitative results}

As motion artifact levels increased, both the PSNR and SSIM values of the motion-corrupted brain images decreased, indicating larger discrepancies relative to the motion-free ground truth. Conversely, the NMSE values increased with higher artifact levels, corroborating the trends observed in PSNR and SSIM. \tablename~\ref{tab:quantitative_values} summarizes the quantitative metrics, and one-way ANOVA tests revealed statistically significant differences across all metrics and distortion levels (all \textit{p} $ \ll 0.0001 $).

Among the compared models, the generative models (Pix2Pix and CycleGAN) significantly (all \textit{p} $ < 0.05 $) improved the quantitative metrics for images with minor and moderate motion artifacts; however, their performance improvements were inconsistent at high artifact levels. In contrast, the U-net model significantly (all \textit{p} $ < 0.05 $) enhanced the metrics across all distortion levels, yielding competitive NMSE values at minor and moderate artifact levels and superior PSNR values at the minor level.

Our proposed PI-MoCoNet achieved superior performance in removing motion artifacts. It produced statistically significantly higher SSIM values across all distortion levels and higher PSNR values at moderate and heavy artifact levels compared to U-net. Furthermore, PI-MoCoNet attained the lowest NMSE values across all distortion levels, with the improvement at the heavy artifact level reaching statistical significance (\textit{p} $ \ll 0.05 $).

\begin{table}[]
	\centering
	\caption{Quantitative Metrics (PSNR [dB], SSIM, NMSE [\%]) across different motion artifact levels.}
	\label{tab:quantitative_values}
	\resizebox{\textwidth}{!}{%
		\begin{tabular}{lllllll}
			\hline
			Metrics                      & Distortion level & Corrupted            & Pix2Pix              & CycleGAN             & U-net                & PI-MoCoNet (ours)    \\ \hline
			\multirow{3}{*}{PNSR [dB]} & Minor            & 34.15 (34.07, 34.24) & 37.03 (36.97, 37.09) & 41.21 (41.11, 41.32) & 44.12 (44.01, 44.24) & 45.95 (45.84, 46.06) \\
			& Moderate         & 30.23 (30.16, 30.30)     & 37.16 (36.98, 37.33) & 38.96 (38.88, 39.04) & 40.36 (40.27, 40.44) & 42.16 (42.08, 42.24) \\
			& Heavy            & 27.99(27.93, 28.06)  & 31.95 (31.84, 32.05) & 35.93 (35.88, 35.98) & 34.53 (34.45, 34.61) & 36.01 (35.94, 36.09) \\ \hline
			\multirow{3}{*}{SSIM [-]}  & Minor            & 0.87 (0.87, 0.88)     & 0.91 (0.91, 0.91) & 0.97 (0.97, 0.97)      & 0.97 (0.97, 0.97)    & 1.00 (1.00, 1.00)    \\
			& Moderate         & 0.80 (0.80, 0.80)    & 0.92 (0.92, 0.93)    & 0.92 (0.91, 0.92)    & 0.96 (0.96, 0.96)    & 0.99 (0.99, 0.99)    \\
			& Heavy            & 0.75 (0.74, 0.75)    & 0.89 (0.88, 0.89)    & 0.91 (0.91, 0.91)    & 0.94 (0.94, 0.94)    & 0.97 (0.97, 0.97)    \\ \hline
			\multirow{3}{*}{NMSE [\%]}  & Minor            & 0.55 (0.54, 0.57)    & 0.15 (0.15, 0.16)    & 0.18 (0.18, 0.19)    & 0.05 (0.05, 0.05)*   & 0.04 (0.04, 0.04)    \\
			& Moderate         & 1.32 (1.30, 1.35)    & 0.57 (0.56, 0.59)    & 0.60 (0.58, 0.62)    & 0.11 (0.11, 0.11)*   & 0.09 (0.09, 0.09)    \\
			& Heavy            & 2.21 (2.17, 2.26)    & 1.14 (1.11, 1.16)    & 0.88 (0.86, 0.90)    & 0.41 (0.40, 0.42)    & 0.36 (0.35, 0.36)    \\ \hline
			\multicolumn{7}{l}{* \textit{p}-value $ > 0.05 $}    \\                                                                                                                          
		\end{tabular}%
	}
\end{table}

Our method could remove motion artifacts from the \textit{in-vivo} MR-ART dataset with two motion  artifact levels. Our method increases PSNR values from 21.23 (95\% CI 21.10, 21.36) and 23.15 (95\% CI 22.98, 23.31) to 31.72 (95\% CI 31.58, 31.85) and 33.01 (95\% CI 32.87, 33.16) dB and SSIM values from 0.72 (95\% CI 0.72, 0.73) and 0.63 (95\% CI 0.63, 0.64) to 0.87 (95\% CI 0.86, 0.87) and 0.83 (95\% CI 0.82, 0.83) for low level and high levels of motion artifacts, respectively. Furthermore, our method reduces NMSE from 10.08 (95\% CI 9.64, 10.55) and 14.77 (95\% CI 14.22, 15.38) to 6.24 (95\% CI 5.96, 6.57) and 8.32 (95\% CI 7.97, 8.72) \% for low and high levels of motion artifact levels, respectively.

\subsection{Ablation study}

To evaluate the contribution of individual loss components in mitigating brain motion artifacts, we conducted an ablation study. Three training scenarios were compared across three levels of motion artifact severity (minor, moderate, and heavy). In the first scenario, the model was trained using only the reconstruction loss, $\mathcal{L}_1$. In the second scenario, the training employed both the reconstruction loss and the data consistency loss, $\mathcal{L}_1 + \mathcal{L}_{\text{dc}}$, thereby excluding the perceptual loss, $\mathcal{L}_{{lpips}}$. Finally, the complete PI-MoCoNet was evaluated, which integrates all three loss terms.
\begin{table}[]
	\centering
	\caption{Ablation study results: Comparison of PSNR [dB], SSIM, and NMSE [\%] (with 95\% confidence intervals) for different loss configurations across motion artifact levels.}
	\label{tab:ablation_study}
	\resizebox{\textwidth}{!}{%
		\begin{tabular}{llll|lll|lll}
			\hline
			\multirow{2}{*}{Scenario} & \multicolumn{3}{c|}{PSNR [dB] (95\% CI)}                                                                                                                                                                           & \multicolumn{3}{c|}{SSIM [-] (95\% CI)}                                                                                                                                                                  & \multicolumn{3}{c}{NMSE [\%] (95\% CI)}                                                                                                                                                                   \\
			& Minor                                                           & Moderate                                                        & Heavy                                                           & Minor                                                        & Moderate                                                     & Heavy                                                        & Minor                                                        & Moderate                                                     & Heavy                                                        \\ \hline
			$ \mathcal{L}_1 $                        & \begin{tabular}[c]{@{}l@{}}46.47\\ (46.37, 46.57)\end{tabular}  & \begin{tabular}[c]{@{}l@{}}40.77\\ (40.69, 40.86\end{tabular}   & \begin{tabular}[c]{@{}l@{}}34.98\\ (34.90, 35.06)\end{tabular}  & \begin{tabular}[c]{@{}l@{}}1.00 \\ (1.00, 1.00)\end{tabular} & \begin{tabular}[c]{@{}l@{}}0.99 \\ (0.99, 0.99)\end{tabular} & \begin{tabular}[c]{@{}l@{}}0.96 \\ (0.96, 0.96)\end{tabular} & \begin{tabular}[c]{@{}l@{}}0.04 \\ (0.04, 0.04)\end{tabular} & \begin{tabular}[c]{@{}l@{}}0.12 \\ (0.12, 0.13)\end{tabular} & \begin{tabular}[c]{@{}l@{}}0.53 \\ (0.51, 0.54)\end{tabular} \\
			$ \mathcal{L}_1  + \mathcal{L}_{\text{dc}}$                    & \begin{tabular}[c]{@{}l@{}}45.92 \\ (45.81, 46.04\end{tabular}  & \begin{tabular}[c]{@{}l@{}}41.77 \\ (41.69, 41.86)\end{tabular} & \begin{tabular}[c]{@{}l@{}}34.52 \\ (34.43, 34.61)\end{tabular} & \begin{tabular}[c]{@{}l@{}}0.99 \\ (0.99, 0.99)\end{tabular} & \begin{tabular}[c]{@{}l@{}}0.99 \\ (0.99, 0.99)\end{tabular} & \begin{tabular}[c]{@{}l@{}}0.97 \\ (0.97, 0.97)\end{tabular} & \begin{tabular}[c]{@{}l@{}}0.04 \\ (0.04, 0.04)\end{tabular} & \begin{tabular}[c]{@{}l@{}}0.10 \\ (0.10, 0.10)\end{tabular} & \begin{tabular}[c]{@{}l@{}}0.46 \\ (0.45, 0.47)\end{tabular} \\
			PI-MoCoNet                & \begin{tabular}[c]{@{}l@{}}45.95 \\ (45.84, 46.06)\end{tabular} & \begin{tabular}[c]{@{}l@{}}42.16 \\ (42.08, 42.24)\end{tabular} & \begin{tabular}[c]{@{}l@{}}36.01 \\ (35.94, 36.09)\end{tabular} & \begin{tabular}[c]{@{}l@{}}1.00 \\ (1.00, 1.00)\end{tabular} & \begin{tabular}[c]{@{}l@{}}0.99 \\ (0.99, 0.99)\end{tabular} & \begin{tabular}[c]{@{}l@{}}0.97 \\ (0.97, 0.97)\end{tabular} & \begin{tabular}[c]{@{}l@{}}0.04 \\ (0.04, 0.04)\end{tabular} & \begin{tabular}[c]{@{}l@{}}0.09 \\ (0.09, 0.09)\end{tabular} & \begin{tabular}[c]{@{}l@{}}0.36 \\ (0.35, 0.36)\end{tabular} \\ \hline
		\end{tabular}%
	}
\end{table}

\tablename~\ref{tab:ablation_study} summarizes the quantitative results—including PSNR, SSIM, and NMSE (with 95\% CIs)—for each scenario. Overall, the SSIM metric remained largely invariant across the different loss configurations and artifact levels, suggesting that SSIM was not substantially influenced by the additional loss terms. In contrast, both PSNR and NMSE benefited from the inclusion of the data consistency and perceptual losses, particularly at higher levels of motion artifacts. Specifically, for heavy artifacts, the full PI-MoCoNet achieved an improvement of approximately 1 dB in PSNR and a reduction of about 0.17 \% in NMSE relative to the $\mathcal{L}_1$-only model.

It is worth noting that although the PSNR for the $\mathcal{L}_1$-only model was occasionally higher than that for the $\mathcal{L}_1 + \mathcal{L}_{\text{dc}}$ variant, the corresponding 95\% confidence interval was noticeably wider, indicating greater variability in performance. Moreover, the complete PI-MoCoNet consistently demonstrated the lowest NMSE values with narrower confidence intervals, underscoring the effectiveness of the combined loss strategy—especially under moderate to heavy motion artifact conditions.

Overall, these results highlight that while minor artifacts are well addressed by even the simplest loss formulation, the integration of data consistency and perceptual losses in PI-MoCoNet is crucial for achieving robust artifact removal at higher distortion levels.

\section{Discussion}\label{sec:discussion}
In this study, we introduced PI-MoCoNet, a novel physics-informed deep learning framework that synergistically leverages both spatial and \textit{k}-space information to mitigate motion artifacts in brain MRI. Unlike conventional methods that either operate solely in the image domain or require explicit estimation of motion parameters, our approach integrates a motion detection network with a motion correction network. This dual-network design, coupled with the incorporation of data consistency and perceptual loss terms, allows PI-MoCoNet to robustly correct for motion-induced artifacts across a wide range of distortion severities.

Our experimental results, obtained on both \textit{in-silico} and \textit{in-vivo} datasets, demonstrate that PI-MoCoNet consistently outperforms established methods such as Pix2Pix GAN, CycleGAN, and a standard U-net architecture. Qualitatively, the method effectively removes ringing and ghosting artifacts even in cases with severe motion corruption, as evidenced by the minimal discrepancies observed in the difference maps relative to the motion-free ground truth. Quantitatively, PI-MoCoNet achieved statistically significant improvements in PSNR, SSIM, and NMSE metrics across all levels of motion artifacts, with particularly notable gains at moderate to heavy distortion levels.

A key strength of the proposed method is its physics-informed design. By enforcing data consistency in \textit{k}-space, the model selectively modifies corrupted regions while preserving the integrity of unaltered areas. This strategy mitigates the risk of image hallucination--a common pitfall in purely image-domain approaches—and ensures that high-fidelity information is maintained throughout the reconstruction process. Moreover, the ablation study underscored the importance of integrating both the data consistency loss and the perceptual loss. While the reconstruction loss alone was adequate for minor motion artifacts, the combined loss formulation was essential for achieving robust performance under more severe conditions, as indicated by the narrower confidence intervals and superior metric values.

Despite these promising results, certain limitations warrant further investigation. First, although our simulated motion artifact framework closely mimics realistic conditions, the performance of PI-MoCoNet should be validated on larger and more diverse clinical datasets. Additionally, while our method is designed for rigid motion artifacts, extending the framework to accommodate non-rigid motions--such as those observed in abdominal imaging--remains an important future direction. Finally, integrating the proposed approach into real-time reconstruction pipelines could further enhance its clinical applicability, reducing the need for repeated scans and improving patient throughput.

Overall, the adoption of a physics-informed deep learning strategy represents a significant advancement in the field of MRI artifact correction. By harnessing domain-specific knowledge of \textit{k}-space properties alongside robust data-driven techniques, PI-MoCoNet offers a promising solution to one of the most persistent challenges in high-resolution brain imaging.

\section{Conclusions}\label{sec:conclusion}

We have presented PI-MoCoNet, a novel framework that effectively addresses brain motion artifacts by combining spatial and \textit{k}-space domain information within a unified deep learning architecture. Through the integration of a motion detection network, a motion correction network, and a comprehensive loss formulation that includes reconstruction, perceptual, and data consistency terms, PI-MoCoNet achieves superior performance over existing methods. Our extensive evaluation on both simulated and \textit{in-vivo} datasets demonstrates statistically significant improvements in PSNR, SSIM, and NMSE, particularly in scenarios with moderate to heavy motion corruption.


\section*{Conflicts of interest}
There are no conflicts of interest declared by the authors.

\section*{Acknowledgment}
This research is supported in part by the National Institutes of Health under Award Numbers R56EB033332, R01EB032680, and R01CA272991.

\section*{Data availability}
The datasets used in this study are publicly available. The IXI dataset can be accessed at \url{https://brain-development.org/ixi-dataset/}, and the MR-ART dataset is available through OpenNeuro~\url{https://openneuro.org/datasets/ds004173/versions/1.0.2}.

%
%

\begin{thebibliography}{10}
	
	\bibitem{Sreekumari217}
	A.~Sreekumari, D.~Shanbhag, D.~Yeo, et~al.,
	\newblock A Deep Learning{\textendash}Based Approach to Reduce Rescan and
	Recall Rates in Clinical MRI Examinations,
	\newblock American Journal of Neuroradiology {\bf 40}, 217--223 (2019).
	
	\bibitem{Villanueva_Meyer2024_Lancet}
	J.~E. Villanueva-Meyer, S.~Bakas, P.~Tiwari, et~al.,
	\newblock Artificial Intelligence for Response Assessment in Neuro Oncology
	(AI-RANO), part 1: review of current advancements,
	\newblock The Lancet Oncology {\bf 25}, e581--e588 (2024).
	
	\bibitem{sharma2024doestimeretreatmentmatter}
	M.~Sharma, I.~E. Naqa, and P.~K. Sneed,
	\newblock Does time to retreatment matter? An NTCP model to predict
	radionecrosis after repeat SRS for recurrent brain metastases incorporating
	time-dependent discounted dose, 2024.
	
	\bibitem{Beigi2018_radiologia}
	M.~Beigi, A.~F. Kazerooni, M.~Safari, et~al.,
	\newblock Heterogeneity analysis of diffusion-weighted MRI for prediction and
	assessment of microstructural changes early after one cycle of induction
	chemotherapy in nasopharyngeal cancer patients,
	\newblock La radiologia medica {\bf 123}, 36--43 (2018).
	
	\bibitem{Beigi2018}
	M.~Beigi, M.~Safari, A.~Ameri, et~al.,
	\newblock Findings of DTI-p maps in comparison with T2/T2-FLAIR to assess
	postoperative hyper-signal abnormal regions in patients with glioblastoma,
	\newblock Cancer Imaging {\bf 18}, 33 (2018).
	
	\bibitem{https://doi.org/10.1002/acm2.14072}
	M.~Safari, A.~Fatemi, Y.~Afkham, and L.~Archambault,
	\newblock Patient-specific geometrical distortion corrections of MRI images
	improve dosimetric planning accuracy of vestibular schwannoma treated with
	gamma knife stereotactic radiosurgery,
	\newblock Journal of Applied Clinical Medical Physics {\bf 24}, e14072 (2023).
	
	\bibitem{https://doi.org/10.1002/jmri.24850}
	M.~Zaitsev, J.~Maclaren, and M.~Herbst,
	\newblock Motion artifacts in MRI: A complex problem with many partial
	solutions,
	\newblock Journal of Magnetic Resonance Imaging {\bf 42}, 887--901 (2015).
	
	\bibitem{9363915}
	S.~K. Zhou, H.~Greenspan, C.~Davatzikos, et~al.,
	\newblock A Review of Deep Learning in Medical Imaging: Imaging Traits,
	Technology Trends, Case Studies With Progress Highlights, and Future
	Promises,
	\newblock Proceedings of the IEEE {\bf 109}, 820--838 (2021).
	
	\bibitem{Zhou2024}
	Z.~Zhou, P.~Hu, and H.~Qi,
	\newblock Stop moving: MR motion correction as an opportunity for artificial
	intelligence,
	\newblock Magnetic Resonance Materials in Physics, Biology and Medicine {\bf
		37}, 397--409 (2024).
	
	\bibitem{DUFFY2021117756}
	B.~A. Duffy, L.~Zhao, F.~Sepehrband, et~al.,
	\newblock Retrospective motion artifact correction of structural MRI images
	using deep learning improves the quality of cortical surface reconstructions,
	\newblock NeuroImage {\bf 230}, 117756 (2021).
	
	\bibitem{Safari_2024_maudgan}
	M.~Safari, X.~Yang, C.-W. Chang, R.~L.~J. Qiu, A.~Fatemi, and L.~Archambault,
	\newblock Unsupervised MRI motion artifact disentanglement: introducing
	MAUDGAN,
	\newblock Physics in Medicine \& Biology {\bf 69}, 115057 (2024).
	
	\bibitem{Liu2021}
	S.~Liu, K.-H. Thung, L.~Qu, W.~Lin, D.~Shen, and P.-T. Yap,
	\newblock Learning MRI artefact removal with unpaired data,
	\newblock Nature Machine Intelligence {\bf 3}, 60--67 (2021).
	
	\bibitem{https://doi.org/10.1002/mp.16844}
	M.~Safari, X.~Yang, A.~Fatemi, and L.~Archambault,
	\newblock MRI motion artifact reduction using a conditional diffusion
	probabilistic model (MAR-CDPM),
	\newblock Medical Physics {\bf 51}, 2598--2610 (2024).
	
	\bibitem{https://doi.org/10.1002/mp.16119}
	J.~Hossbach, D.~N. Splitthoff, S.~Cauley, et~al.,
	\newblock Deep learning-based motion quantification from k-space for fast
	model-based magnetic resonance imaging motion correction,
	\newblock Medical Physics {\bf 50}, 2148--2161 (2023).
	
	\bibitem{10639524}
	O.~Dabrowski, J.-L. Falcone, A.~Klauser, et~al.,
	\newblock SISMIK for Brain MRI: Deep-Learning-Based Motion Estimation and
	Model-Based Motion Correction in k-Space,
	\newblock IEEE Transactions on Medical Imaging {\bf 44}, 396--408 (2025).
	
	\bibitem{10681246}
	Z.~Yue, J.~Wang, and C.~C. Loy,
	\newblock Efficient Diffusion Model for Image Restoration by Residual Shifting,
	\newblock IEEE Transactions on Pattern Analysis and Machine Intelligence {\bf
		47}, 116--130 (2025).
	
	\bibitem{doi:https://doi.org/10.1002/9781118633953.ch10}
	{\em Magnetic Resonance Imaging}, chapter~10, pages 165--206,
	\newblock John Wiley \& Sons, Ltd, 2014.
	
	\bibitem{9025260}
	R.~Shaw, C.~H. Sudre, T.~Varsavsky, S.~Ourselin, and M.~J. Cardoso,
	\newblock A k-Space Model of Movement Artefacts: Application to Segmentation
	Augmentation and Artefact Removal,
	\newblock IEEE Transactions on Medical Imaging {\bf 39}, 2881--2892 (2020).
	
	\bibitem{10643318}
	R.~Azad, E.~K. Aghdam, A.~Rauland, et~al.,
	\newblock Medical Image Segmentation Review: The Success of U-Net,
	\newblock IEEE Transactions on Pattern Analysis and Machine Intelligence {\bf
		46}, 10076--10095 (2024).
	
	\bibitem{Li_ASPS_MICCAI2024}
	H.~Li, D.~Zhang, J.~Yao, L.~Han, Z.~Li, and J.~Han,
	\newblock { ASPS: Augmented Segment Anything Model for Polyp Segmentation },
	\newblock in {\em proceedings of Medical Image Computing and Computer Assisted
		Intervention -- MICCAI 2024}, volume LNCS 15009, Springer Nature Switzerland,
	2024.
	
	\bibitem{10852524}
	A.~Li, L.~Zhang, Y.~Liu, and C.~Zhu,
	\newblock Exploring Frequency-Inspired Optimization in Transformer for
	Efficient Single Image Super-Resolution,
	\newblock IEEE Transactions on Pattern Analysis and Machine Intelligence ,
	1--18 (2025).
	
	\bibitem{https://doi.org/10.1002/mp.17600}
	Z.~Eidex, M.~Safari, R.~L.~J. Qiu, et~al.,
	\newblock T1-contrast enhanced MRI generation from multi-parametric MRI for
	glioma patients with latent tumor conditioning,
	\newblock Medical Physics {\bf n/a}.
	
	\bibitem{9607618}
	J.~Liang, J.~Cao, G.~Sun, K.~Zhang, L.~Van~Gool, and R.~Timofte,
	\newblock SwinIR: Image Restoration Using Swin Transformer,
	\newblock in {\em 2021 IEEE/CVF International Conference on Computer Vision
		Workshops (ICCVW)}, pages 1833--1844, 2021.
	
	\bibitem{8578166}
	R.~Zhang, P.~Isola, A.~A. Efros, E.~Shechtman, and O.~Wang,
	\newblock The Unreasonable Effectiveness of Deep Features as a Perceptual
	Metric,
	\newblock in {\em 2018 IEEE/CVF Conference on Computer Vision and Pattern
		Recognition}, pages 586--595, 2018.
	
	\bibitem{paszke2019pytorchimperativestylehighperformance}
	A.~Paszke, S.~Gross, F.~Massa, et~al.,
	\newblock PyTorch: An Imperative Style, High-Performance Deep Learning Library,
	2019.
	
	\bibitem{Narai2022_mrART}
	{\'A}.~N{\'a}rai, P.~Hermann, T.~Auer, et~al.,
	\newblock Movement-related artefacts (MR-ART) dataset of matched
	motion-corrupted and clean structural MRI brain scans,
	\newblock Scientific Data {\bf 9}, 630 (2022).
	
	\bibitem{1284395}
	Z.~Wang, A.~Bovik, H.~Sheikh, and E.~Simoncelli,
	\newblock Image quality assessment: from error visibility to structural
	similarity,
	\newblock IEEE Transactions on Image Processing {\bf 13}, 600--612 (2004).
	
	\bibitem{kastryulin2022pytorchimagequalitymetrics}
	S.~Kastryulin, J.~Zakirov, D.~Prokopenko, and D.~V. Dylov,
	\newblock PyTorch Image Quality: Metrics for Image Quality Assessment, 2022.
	
	\bibitem{safari2024selfsupervisedadversarialdiffusionmodels}
	M.~Safari, Z.~Eidex, S.~Pan, R.~L.~J. Qiu, and X.~Yang,
	\newblock Self-Supervised Adversarial Diffusion Models for Fast MRI
	Reconstruction, 2024.
	
	\bibitem{seabold2010statsmodels}
	S.~Seabold and J.~Perktold,
	\newblock Statsmodels: econometric and statistical modeling with python.,
	\newblock SciPy {\bf 7} (2010).
	
\end{thebibliography}

\end{document}